\documentclass[utf8]{frontiersSCNS} 

\usepackage{url,hyperref,lineno,microtype,subcaption}
\usepackage[onehalfspacing]{setspace}

\def\keyFont{\fontsize{8}{11}\helveticabold }
\def\firstAuthorLast{Ramcharan {et~al.}} 
\def\Authors{Amanda Ramcharan\,$^{1}$, Kelsee Baranowski\,$^{1}$, Peter McCloskey\,$^{2}$,Babuali Ahmed\,$^{3}$, James Legg\,$^{3}$, and David Hughes\,$^{1,4,5*}$}
\def\Address{$^{1}$Department of Entomology, College of Agricultural Sciences, Penn State University, State College, PA, USA \\
$^{2}$Department of Computer Science, Pittsburgh University,Pittsburgh, PA, USA  \\
$^{3}$International Institute for Tropical Agriculture, Dar el Salaam, Tanzania \\
$^{4}$Department of Biology, Eberly College of Sciences, Penn State University, State College, PA, USA  \\ 
$^{5}$Center for Infectious Disease Dynamics, Huck Institutes of Life Sciences, Penn State University, State College, PA, USA }
\def\corrAuthor{David Hughes}

\def\corrEmail{dph14@psu.edu}

\begin{document}
\onecolumn
\firstpage{1}

\title[Transfer Learning for Cassava Disease Detection]{Using Transfer Learning for Image-Based Cassava Disease Detection} 

\author[\firstAuthorLast ]{\Authors} 
\address{} 
\correspondance{} 

\extraAuth{}

\maketitle

\begin{abstract}

\section{}

Cassava is the third largest source of carbohydrates for human food in the world but is vulnerable to virus diseases, which threaten to destabilize food security in sub-Saharan Africa. Novel methods of cassava disease detection are needed to support improved control which will prevent this crisis. Image recognition offers both a cost effective and scalable technology for disease detection. New transfer learning methods offer an avenue for this technology to be easily deployed on mobile devices. Using a dataset of cassava disease images taken in the field in Tanzania, we applied transfer learning to train a deep convolutional neural network to identify three diseases and two types of pest damage (or lack thereof). The best trained model accuracies were 98\% for brown leaf spot (BLS), 96\% for red mite damage (RMD), 95\% for green mite damage (GMD), 98\% for cassava brown streak disease (CBSD), and 96\% for cassava mosaic disease (CMD).  The best model achieved an overall accuracy of 93\% for data not used in the training process. Our results show that the transfer learning approach for image recognition of field images offers a fast, affordable, and easily deployable strategy for digital plant disease detection.\\

\tiny
 \keyFont{ \section{Keywords:} cassava disease detection, convolutional neural networks, transfer learning, mobile epidemiology, Inception v3 model} 
\end{abstract}

\section{Introduction}
Cassava (\textit{Manihot esculenta} Crantz) is the most widely grown root crop in the world and a major source of calories for roughly two out of every five Africans \cite{nweke2002cassava}. In 2014, over 145 million tonnes of cassava were harvested on 17 million hectares of land on the African continent \cite{FAOSTAT}. It is considered a food security crop for smallholder farms, especially in low-income, food-deficit areas \cite{bellotti1999recent} as it provides sufficient yields in low soil fertility conditions and where there are irregular rainfall patterns \cite{de1989importance}.

Smallholder farmers, representing 85\% of the world's farms, face numerous risks to their agricultural production such as climate change, market shocks, and pest and disease outbreaks \cite{nagayet2005future}. Cassava, an exotic species introduced to Africa from South America in the 16th century, initially had few pest and disease constraints on the continent. In the 1970s two arthropod pests, the cassava mealybug (\textit{Phenacoccus manihoti}(Matt.-Ferr.)) and the cassava green mite (\textit{Mononychellus tanajoa} (Bond.)) were accidentally introduced from the neotropics \cite{legg1999emergence}, becoming the most economically threatening pests. Cassava virus diseases, in particular cassava mosaic disease (CMD) and cassava brown streak disease (CBSD), have a longer history on the continent. CMD was the first to be recorded in Tanzania towards the end of the 19th century \cite{warburg1984}. In East Africa, the outbreak of a severe form of the virus in the 1990s, termed EACMV-UG (or UgV), coupled with the sensitivity of local cultivars, resulted in a threat to food security in the region as farmers’ only solution was to abandon cultivation \cite{thresh1994effects}. Thresh et al. (1997) estimated annual yield losses to CMD at 15-24\%, or 21.8-34.8 million tonnes, at 1994 production levels. CBSD was reported in the 1930s \cite{storey1936virus}. With limited success in controlling CMD and CBSD, the two diseases have become the largest constraints to cassava production and food security in sub-Saharan Africa resulting in losses of over US\$1 billion every year \cite{legg2006cassava}.

In order to manage the detection and spread of cassava diseases, early identification in the field is a crucial first step. Traditional disease identification approaches rely on the support of agricultural extension organizations, but these approaches are limited in countries with low logistical and human infrastructure capacity, and are expensive to scale up. In such areas, internet penetration, smartphone and unmanned aerial vehicle (UAV) technologies offer new tools for in-field plant disease detection based on automated image recognition that can aid in early detection at a large scale. Previous research has demonstrated automated image recognition of crop diseases in wheat \cite{gibson2015towards}, \cite{siricharoen2016lightweight}, apples \cite{dubey2014adapted} and on datasets of healthy and diseased plants \cite{mohanty2016using}; this technology was also demonstrated on UAVs \cite{puig2015assessment}. Cassava disease detection based on automated image recognition through feature extraction has shown promising results \cite{aduwo2010automated}, \cite{abdullakasim2011images}, \cite{mwebaze2016machine} but extracting features is computationally intensive and requires expert knowledge for robust performance. In order to capitalize on smartphone technology, models must be fast and adapted to limited processing power. Transfer learning, where a model that has been trained on a large image dataset is retrained for new classes, offers a shortcut to training image recognition models because of lower computational requirements. This would have a distinct advantage for field settings.  Here we investigated the potential for adapting an already trained deep learning convolutional neural network model to detect incidence of cassava disease using an in-field dataset of 2,756 images comprising 3 cassava diseases and 2 types of pest damage (or lack thereof).

\section{Methods}
\subsection{The Cassava Image Datasets}
\begin{figure}
\begin{center}
\includegraphics[width=15cm]{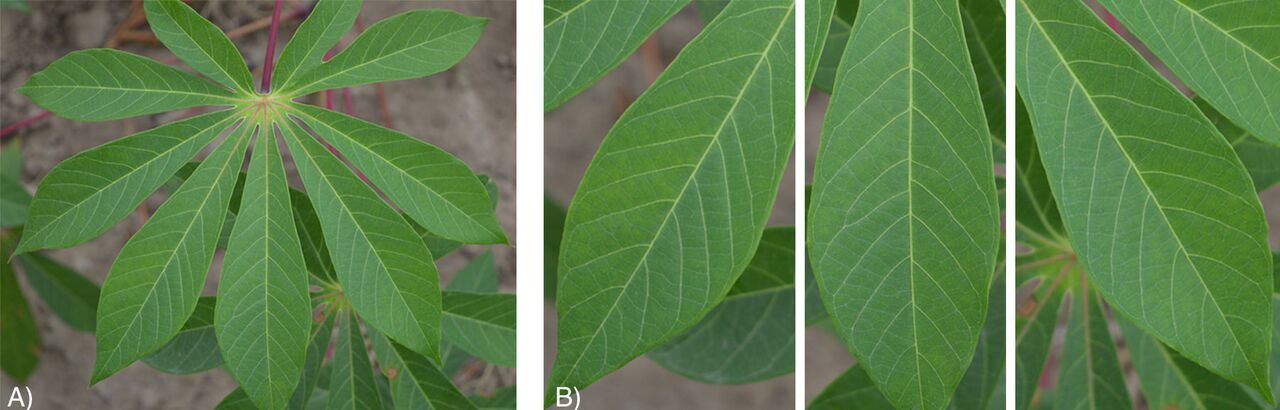}
\end{center}
\caption{Healthy cassava leaf images from the A) original cassava dataset and B) leaflet dataset}\label{fig:1}
\end{figure}

The cassava leaf images were taken with a commonly available Sony Cybershot 20.2-megapixel digital camera in experimental fields belonging to the International Institute for Tropical Agriculture (IITA), outside of Bagamoyo, Tanzania.  The entire cassava leaf roughly centered in the frame was photographed to build the first dataset. Over a four-week period, 11,670 images were taken. Images were then screened for co-infections to limit the number of images with multiple diseases. This dataset, called the ‘original cassava dataset’, comprised 2,756 images. These photos were then manually cropped into individual leaflets to build the second dataset. This dataset, called the ‘leaflet cassava dataset’, comprised 15,000 images of cassava leaflets (2,500 images per class). Figure \ref{fig:1} shows an example from both datasets: a) original cassava dataset and b) leaflet cassava dataset. Both datasets were tested to shed light on model performance with images of full leaves but fewer images versus cropped leaves with more images. The underlying assumption was that the cropped leaf images (leaflet cassava dataset) would improve model performance to correctly identify a disease as the dataset was larger. Additionally, we suspected the end users trying to get a diagnosis for a disease would focus in on leaflets showing symptoms. Both datasets comprised six class labels assigned manually based on in-field diagnoses by cassava disease experts from IITA. For all datasets, we used the images as is, in color, and with varied backgrounds from the field in order to assess the model performance (Figure S6).

\begin{figure}
\begin{center}
\includegraphics[width=15cm]{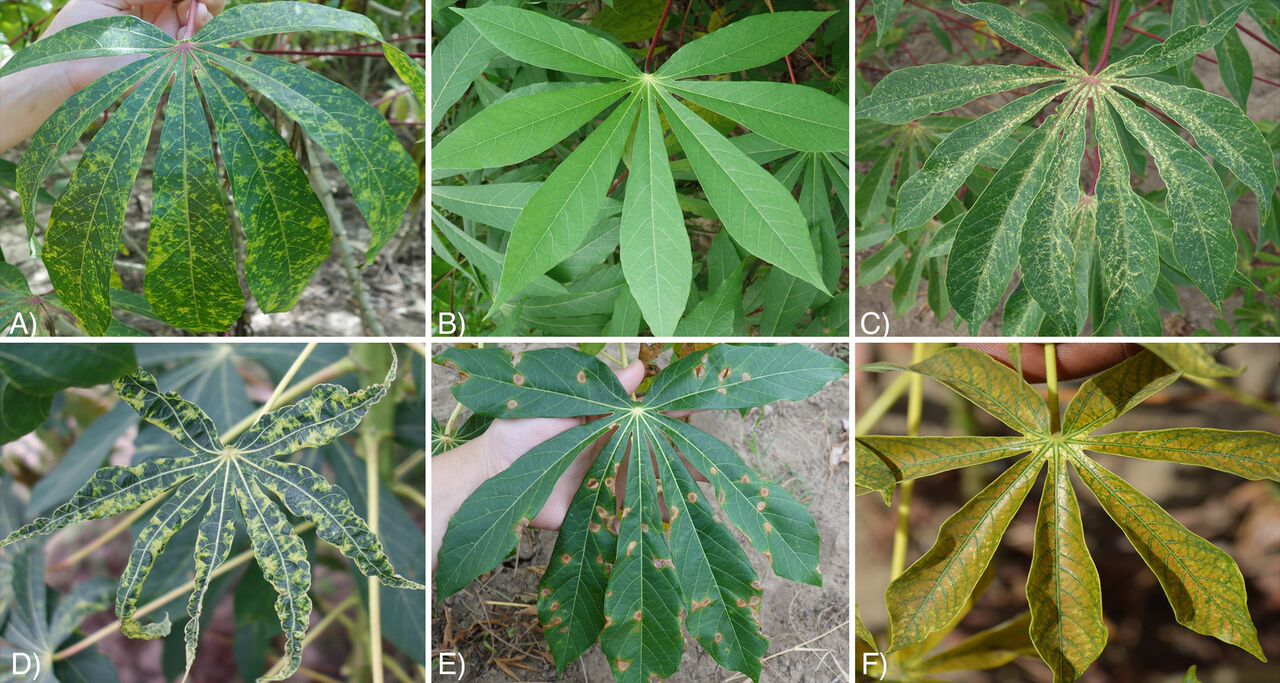}
\end{center}
\caption{Examples images with in field backgrounds from 6 classes in the original cassava dataset. A) Cassava Brown Streak Virus (CBSD), B) Healthy,M C) Green Mite Damage (GMD), D) Cassava Mosaic Virus (CMD), E) Cassava Brown Leaf Spot (CBLS), F) Red Mite Damage (RMD).
}\label{fig:2}
\end{figure}

The six class labels for the datasets included three disease classes, two mite damage classes, and one healthy class, defined as a lack of disease or mite damage on the leaf. The disease classes and the number of images in the original dataset were: CBSD (398 images), CMD (388 images), brown leaf spot (BLS) (386 images), and the mite damage classes were: cassava green mite damage (GMD) (309 images) and red mite damage (RMD) (415 images). Figure \ref{fig:2} illustrates examples of the class labels for the original cassava dataset. Within these classes, several cassava varieties were photographed at different stages of plant maturity (Table S1).

The five disease and pest class symptoms are as follows:

CBSD is a result of infection with cassava brown streak ipomoviruses (CBSIs) (family \textit{Potyviridae}, genus \textit{Ipomovirus}). There are two species associated with the disease, cassava brown streak virus (CBSV) and Ugandan cassava brown streak virus (UCBSV) \cite{mbanzibwa2011simultaneous}.  Both cause the same symptoms. These two virus species are vectored by whiteflies (\textit{Bemisia tabaci} (Genn.)) in a semi-persistent manner. When infected, cassava leaves show a mottled yellowing pattern typically beginning from the secondary veins and progressing to tertiary veins as the infection gets more severe \cite{nichols1950brown}. This yellowish chlorosis spreads along the veins until severely infected leaves are mostly yellow. Disease symptoms can vary by variety, age of the plant and weather conditions. Tolerant varieties and plants at a young age may be infected but asymptomatic. The two viruses may also cause brown streaks on stems of infected plants and brown necrotic rotting in tuberous roots which may render them inedible. 

CMD is a result of infection with cassava mosaic begomoviruses (CMBs) (family \textit{Geminiviridae}, genus \textit{Begomovirus}). There are many species and recombinant strains associated with this group of viruses, although the common form in coastal East Africa, where sampling was undertaken, is \textit{East African cassava mosaic virus} (EACMV) \cite{ndunguru2005molecular}. The virus species are vectored by \textit{B. tabaci} (Genn.) in a persistent manner, contrasting to CBSIs. Newly-infected plants begin to express symptoms from the top, while plants infected through the planted cutting often show symptoms in all leaves. Symptoms of CMD are a typical mosaic in which there is a mix of yellow/pale green chlorotic patches and green areas (Figure \ref{fig:2}d). Unlike CBSD, leaves are usually distorted in shape, and where symptoms are severe the size of leaves is greatly reduced and the plant is stunted. Stunting and the damage to chlorophyll resulting from chlorosis results in the quantitative declines in yield.

BLS is caused by the fungus (family \textit{Mycosphaerella} genus \textit{henningsii} (Sivan)). This fungus is distributed worldwide and typically does not cause great yield loss. The disease manifests in brown circular leaf spots with some varieties expressing a chlorotic halo around the spots. Severe infections can cause the leaves to turn yellow or brown. The circular spots can become dry and crack depending on the environment.  

GMD is caused by cassava green mites (\textit{Mononychellus tanajo} (Bondar)). This is a widespread pest in Africa and South America. The mites cause small white, scratch-like spots where they have fed and in severe cases cause the whole leaf to be covered with the pattern. There is such a reduction in chlorophyll that the leaf may become stunted in a manner similar to that caused by CMD. Depending on variety and environment, infestations can lead to losses in tuberous root yield of up to 30\% \cite{Skovgard1993}.

RMD is caused by cassava red spider mite (\textit{Oligonychus biharensis} (Hirst)), which is widely distributed across Africa. Their feeding damage also causes small scratch-like spotting on the leaf but typically produces a distinct reddish-brown rust color. Feeding is also focused around the main vein but severe infestations can cause the whole leaf to turn orange.

Although GMD and RMD are not strictly diseases, for simplicity, we refer to all of the conditions affecting the plants that were considered in this study as ‘diseases’.

\subsection{Approach}
We evaluated the applicability of transfer learning from a deep convolutional neural network model for the cassava image datasets. Using the latest version of the successful Inception model (based on GoogLeNet), we implemented the image recognition code for Inception v3 in TensorFlow. Inception v3 is trained from the ImageNet Large Visual Recognition Challenge using the data from 2012, where it was tasked with classifying images into 1,000 classes. The top-5 error rate of Inception v3 was 3.46\%, compared to 6.67\% and 4.9\% for Inception (GoogLeNet) and BN-Inception v2 respectively \cite{szegedy2016rethinking}. Inception v3 is 42 layers deep, but the computation cost is only 2.5 times higher than that of GoogLeNet with 22 layers. Beginning with the GoogLeNet model, Inception v3 implements several design principles to scale up convolutional networks to improve performance with a modest increase in computational cost; a significant benefit for scenarios where memory or computational power is limited, such as mobile or drone devices. Beginning the the GoogLeNet model, Inception v3 factorizes the traditional 7 x 7 convolution into three 3 x 3 convolutions, grid reduction is applied to three traditional inception modules to reduce to a 17 x 17 grid with 768 filters, then grid reduction is applied again to five factorized inception modules to reduce to a 8 x 8 x 1280 grid. A detailed description of the design principles implemented to create the Inception v3 model from GoogleNet is provided in \cite{szegedy2016rethinking}. The model parameters implemented in this study included the number of training steps (4000), the learning rate (0.035), train batch size (100), test batch size (-1; the entire test set), and the validation batch size (100).

Transfer learning retrains the final layer of the Inception v3 model to classify a new dataset by exploiting the large amount of visual knowledge already learned from the Imagenet database. Previous research has shown that transfer learning is effective for many applications \cite{mohanty2016using}, \cite{karpathy2014large}, \cite{yosinski2014transferable} and has much lower computational requirements than learning from scratch. We analyzed the performance of training the final layer of Inception v3 with three different architectures: the original inception softmax layer, support vector machines (SVM), and knn nearest neighbor(knn). 

\subsection{Performance Measurements}
In order to perform a robust validation and test for any inherent bias in the datasets, experiments were run for a range of training-testing data splits. During model training, 10\% of the dataset was used to validate training steps, thus 90\% of the dataset was split into different training and testing dataset configurations. The training-test splits were as follows: 80-10 (80\% of dataset for training, 10\% for testing respectively), 60-30 (60\% of dataset for training, 30\% for testing respectively), 50-40, (50\% of dataset for training, 40\% for testing respectively), 40-50 (40\% of dataset for training, 50\% for testing respectively), and 20-70 (20\% of dataset for training, 70\% for testing respectively). For each experiment the overall accuracy is reported as the number of samples in all classes were similar.

\section{Results}
\begin{figure}
\begin{center}
\includegraphics[width=15cm]{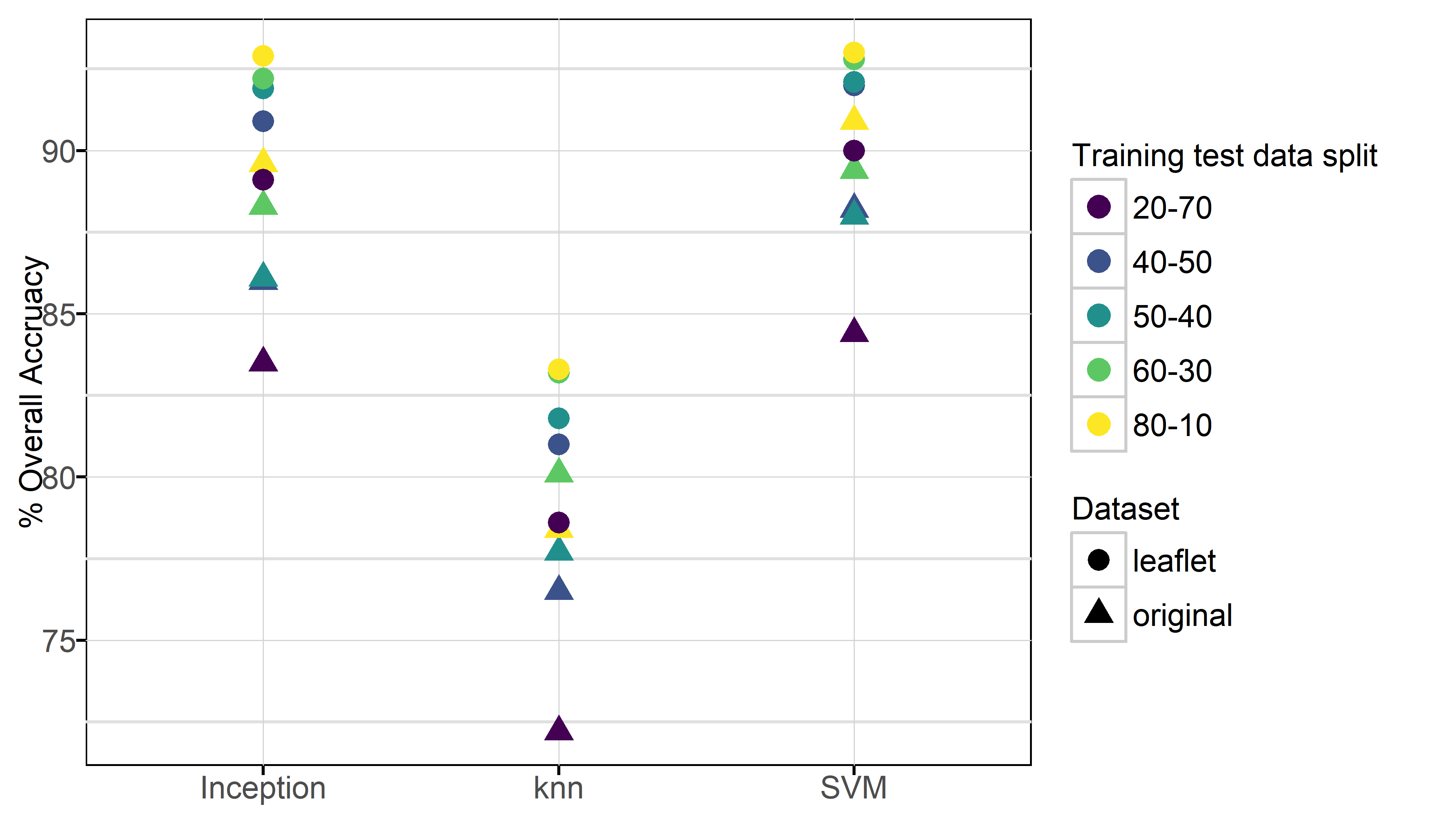}
\end{center}
\caption{Overall accuracy for transfer learning using three machine learning methods}\label{fig:3}
\end{figure}

For the original cassava dataset (i.e. the whole leaf), the overall accuracy in classifying a leaf as belonging to the correct category ranged from 73\% (20-70 split, knn) to 91\% (80-10 split, SVM). For the leaflet cassava dataset, the overall accuracy ranged from was higher and ranged from 80\% (20-70 split, knn) to 93.0\% (80-10 split, SVM). Figure \ref{fig:3} and Table S2 show the overall accuracies for the datasets. It is worth noting that all models performed much better than randomly guessing, even with varied backgrounds in the images such as human hands, feet, soil or other distracting features. Results also suggest the models were not overfit to the datasets as the training-testing data split had a small effect on the overall accuracies reported. Support vector machines and the final Inception v3 softmax layer, both based on achieving linear separability of the classes, had similar model performances for both original and leaflet datasets, while the knn model (k=3), based on similarity with its neighbors, performed the worst.

The confusion matrices from the original and leaflet dataset, allow a more detailed analysis by shedding light on how model performance changes with different disease representations in the images. On the confusion matrix plots for the 80-10 (80\% of dataset for training, 10\% for testing respectively) data split, the rows correspond to the true class, and the columns show the predicted class. The diagonal cells show the proportion (range 0-1) of the examples the trained network correctly predicts the classes of observations i.e. this is the proportion of the examples in which the true and predicted classes match. The off-diagonal cells show where the network made mistakes. 

\begin{figure}
\begin{center}
\includegraphics[width=14.5cm]{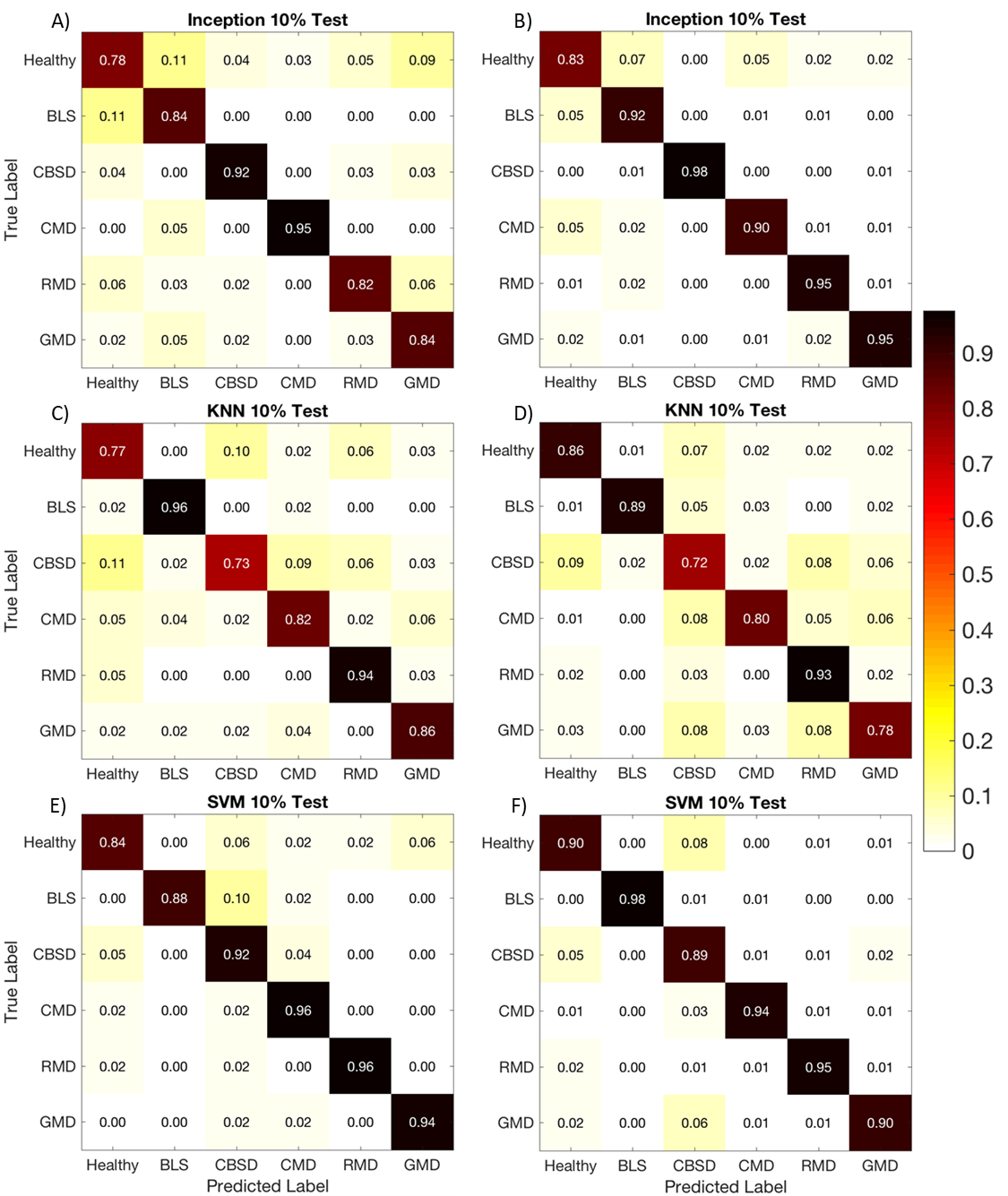}
\end{center}
\caption{Confusion matrices for 10\% test dataset using original (A, C, E) and leaflet (B,D,F) datasets.}\label{fig:4}
\end{figure}

Using the confusion matrix for the Inception v3 model in Figure \ref{fig:4}a (original dataset, 80-10 data split) as an example, the Healthy class diagonal cell shows the Inception v3 model correctly identified 0.78 or 78\% of the healthy leaf images. This increased to 0.83 for the leaflet data set (Figure \ref{fig:4}b). Overall the target and output class predictions were within a similar range for all classes and datasets suggesting the overall reported accuracy is indicative of the model performance for all cassava disease classes. 

Comparing proportions in the off-diagonal cells for the models showed that the proportion of correct predictions between the leaflet dataset and the original dataset did not significantly increase even though the leaflet dataset was almost 7 times as large as the original dataset. Looking into specific diseases in Figure \ref{fig:4}, the highest reported prediction accuracy was 0.98 for CBSD (Inception v3-leaflet) and BLS (SVM- leaflet) diseases. The Inception v3 model had the highest accuracies of 0.98 and 0.95 with the leaflet dataset for CBSD and GMD, respectively. Healthy and BLS classes also had highest accuracies with the leaflet dataset for the SVM model (0.90 and 0.98 respectively). The slight improvement in accuracies using the leaflet dataset over the original dataset could be due to the increase in sample size for the disease classes providing more images for the models to learn from. Alternatively, the leaflet dataset could reduce the accuracy of the model as all leaflets on a cassava leaf may not show signs of a disease, which would confuse the model.  For CMD and RMD, the SVM model with the original dataset had the highest accuracies of 0.96 and 0.96, respectively. These results suggest the size of the dataset is not as important in improving prediction accuracy as previously assumed. 

\section{Discussion and Conclusion}
The results of this study show that image recognition with transfer learning from the convolutional neural network Inception v3 is a powerful method for automated cassava disease detection. This method avoids the complex and labor-intensive step of feature extraction from images in order to train models. Transfer learning is also capable of applying common machine learning methods by retraining the vectors produced by the trained model on new class data. In this study, three machine learning methods were used, and results showed the SVM model to be have the highest prediction accuracies for four out of six disease classes. With respect to specific cassava diseases, the SVM model had the highest accuracies for cassava mosaic disease (CMD) (96\%) and red mite damage (RMD) (96\%) using the original dataset, and 90\% and 98\% for Healthy and brown leaf spot (BLS) using the leaflet dataset. The Inception v3 model had the highest accuracies for the cassava brown streak virus (CBSD) (98\%) and 95\% accuracy for green mite damage (GMD) with the leaflet dataset. 

In a practical field setting where the goal is smartphone assisted disease diagnosis, our results show that diagnostic accuracy improved only slightly when the leaflet was used rather than the whole leaf for some diseases (CBSD, BLS, GMD), while whole leaf images gave higher accuracies for other diseases (CMD and RMD). This was not expected. Rather, the larger image leaflet dataset was expected to perform better for all disease classes compared to the original dataset. These results indicate that datasets needed to build transfer learning models for plant disease diagnosis do not require very large training datasets (< 500 images per class). The high accuracies reported suggest that variations in background had little effect on the prediction accuracies of the model. Portions of images contained the sky, hands, shoes, and other vegetation, yet predictions in all image classes were above the probability of randomly guessing (16.7\%). In the field it is also likely that an extension worker would use more than one picture to predict the disease, thus improving the diagnostic accuracy further.  This study therefore shows that transfer learning offers a promising avenue for in-field disease detection using convolutional neural networks with relatively small image datasets. Future work to validate the method in the field should aim to develop mobile phone-based applications that would allow technicians to rapidly monitor disease prevalence.

\section*{Conflict of Interest Statement}
The authors declare that the research was conducted in the absence of any commercial or financial relationships that could be construed as a potential conflict of interest.

\section*{Author Contributions}
AR, KB,BA, JL and DH conceived the study and wrote the paper.BA, KB, and JL collected and processed data. PM and AR implemented the algorithms described and prepared results. 

\section*{Funding}
We thank the Huck Institutes at Penn State University for support.

\bibliographystyle{frontiersinSCNS_ENG_HUMS} 
\bibliography{test}


\renewcommand{\arraystretch}{1.5}



\end{document}